\pdfoutput=1
\documentclass[11pt]{article}

\usepackage{emnlp2021}

\usepackage{times}
\usepackage{latexsym}

\usepackage[T1]{fontenc}
\usepackage[utf8]{inputenc}

\usepackage{microtype}
\usepackage{hyperref}
\usepackage{url}
\usepackage{booktabs}
\usepackage{multirow}
\usepackage{graphicx}
\usepackage{xspace}
\usepackage{subcaption}
\usepackage{amsmath}
\usepackage{xcolor}
\usepackage[export]{adjustbox}
\newcommand{\task}{\textsc{XFUND}\xspace}

\title{LayoutXLM: Multimodal Pre-training for \\Multilingual Visually-rich Document Understanding}
\author{Yiheng Xu$^{1}$\thanks{Work done during internship at Microsoft Research Asia.}, Tengchao Lv$^{1}$,  Lei Cui$^{1}$, Guoxin Wang$^{2}$, Yijuan Lu$^{2}$, \\ \textbf{Dinei Florencio$^{2}$, Cha Zhang$^{2}$, Furu Wei$^{1}$} \\
$^{1}$Microsoft Research Asia\\
$^{2}$Microsoft Azure AI\\
\texttt{\{t-yihengxu,tengchaolv,lecu,guow,yijlu,dinei,chazhang,fuwei\}@microsoft.com} \\
}

\begin{document}
\maketitle
\begin{abstract}
	% This document is a supplement to the general instructions for *ACL authors. It contains instructions for using the \LaTeX{} style files for EMNLP 2021. 
	% The document itself conforms to its own specifications, and is therefore an example of what your manuscript should look like.
	% These instructions should be used both for papers submitted for review and for final versions of accepted papers.

	Multimodal pre-training with text, layout, and image has achieved SOTA performance for visually-rich document understanding tasks recently, which demonstrates the great potential for joint learning across different modalities. In this paper, we present \textbf{LayoutXLM}, a multimodal pre-trained model for multilingual document understanding, which aims to bridge the language barriers for visually-rich document understanding. To accurately evaluate LayoutXLM, we also introduce a multilingual form understanding benchmark dataset named \textbf{\task}, which includes form understanding samples in 7 languages (Chinese, Japanese, Spanish, French, Italian, German, Portuguese), and key-value pairs are manually labeled for each language. Experiment results show that the LayoutXLM model has significantly outperformed the existing SOTA cross-lingual pre-trained models on the \task dataset. The pre-trained LayoutXLM model and the \task dataset are publicly available at \url{https://aka.ms/layoutxlm}.

\end{abstract}

\section{Introduction}

Multimodal pre-training for visually-rich Document Understanding (VrDU) has achieved new SOTA performance on several public benchmarks recently~\citep{xu2020layoutlmv2,10.1145/3394486.3403172}, including form understanding~\citep{Jaume_2019}, receipt understanding~\citep{park2019cord}, complex layout understanding~\citep{graliski2020kleister}, document image classification~\citep{harley2015icdar} and document VQA task~\cite{mathew2020docvqa}, due to the advantage that text, layout and image information is jointly learned end-to-end in a single framework. Meanwhile, we are well aware of the demand from the non-English world since nearly 40\% of digital documents on the web are in non-English languages. Simply translating these documents automatically with machine translation services might help, but it is often not satisfactory due to the poor translation quality on document images~\citep{afli-way-2016-integrating}. Therefore, it is vital to pre-train the LayoutLM model using real document datasets around the world for the multilingual VrDU task.

\begin{figure*}[t]
  \centering
  \includegraphics[width=1\textwidth]{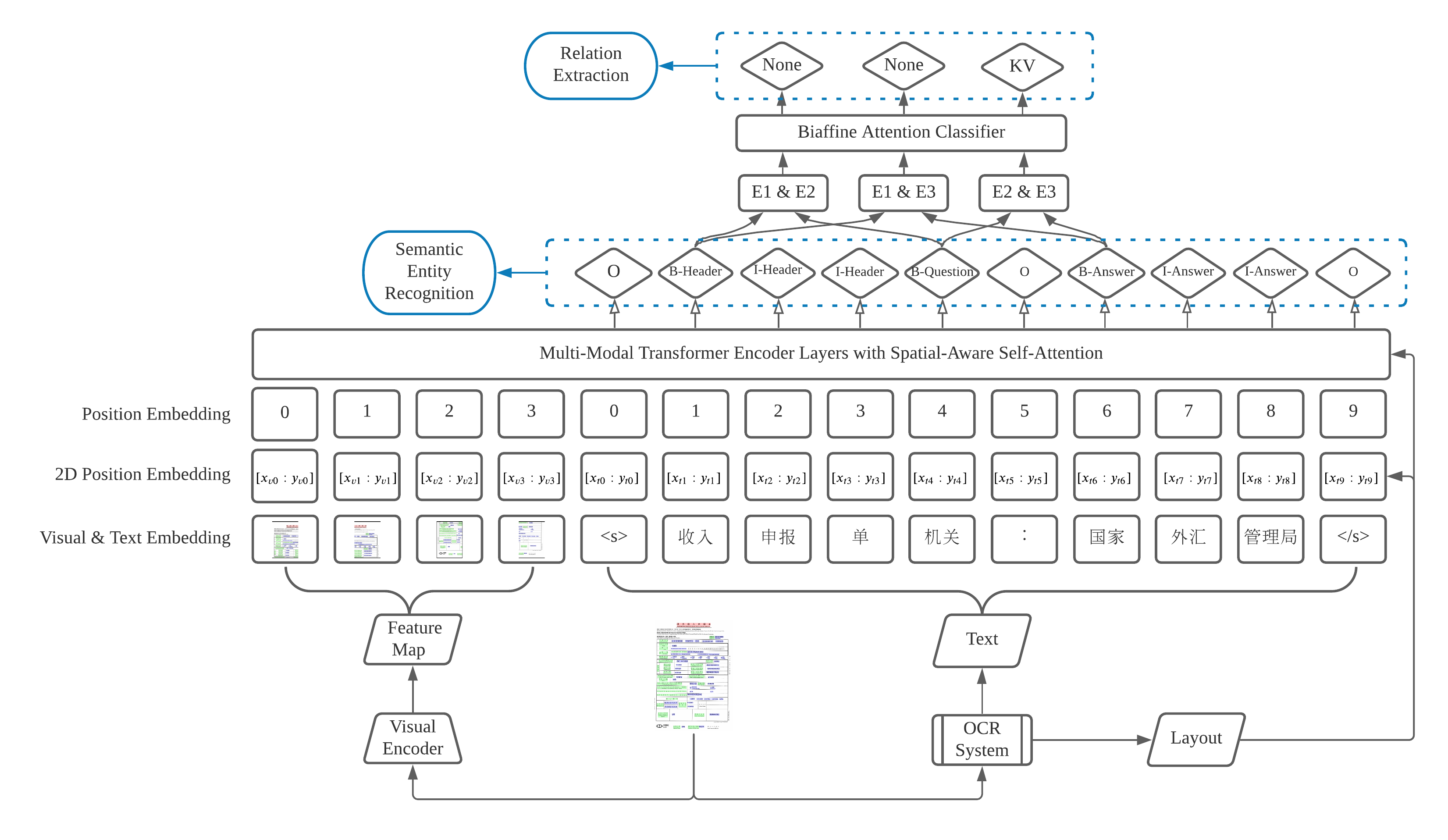}
  \caption{Architecture of the LayoutXLM Model, where the semantic entity recognition and relation extraction tasks are also demonstrated.}
  \label{fig:arch}
\end{figure*}

Multilingual pre-trained models such as mBERT~\citep{devlin2018bert}, XLM~\citep{lample2019crosslingual}, XLM-RoBERTa~\citep{conneau2020unsupervised}, mBART~\citep{liu2020multilingual}, and the recent InfoXLM~\citep{chi2020infoxlm} and mT5~\citep{xue2020mt5} have pushed many SOTA results on cross-lingual natural language understanding tasks by pre-training the Transformer models on different languages. These models have successfully bridged the language barriers in a number of cross-lingual transfer benchmarks such as XNLI~\citep{conneau2018xnli} and XTREME~\citep{hu2020xtreme}. Although a large amount of multilingual text data has been used in these cross-lingual pre-trained models, text-only multilingual models cannot be easily used in the VrDU tasks because they are usually fragile in analyzing the documents due to the format/layout diversity of documents in different countries, and even different regions in the same country. Hence, to accurately understand these visually-rich documents in various languages, it is crucial to pre-train the multilingual models with not only textual information but also layout and image information in a multimodal framework.

To this end, we present a multimodal pre-trained model for multilingual VrDU tasks in this paper, aka \textbf{LayoutXLM}, which is a multilingual extension of the recent LayoutLMv2 model~\citep{xu2020layoutlmv2}. LayoutLMv2 integrates the image information in the pre-training stage by taking advantage of the Transformer architecture to learn the cross-modality interaction between visual and textual information.
In addition, LayoutLMv2 uses two new training objectives in addition to the masked visual-language model, which are the image-text matching and image masking prediction tasks. In this way, the pre-trained models absorb cross-modal knowledge from different document types, where the local invariance among the layout and formats is preserved. Inspired by the LayoutLMv2 model, LayoutXLM adopts the same architecture for the multimodal pre-training initialized by a SOTA multilingual pre-trained InfoXLM model~\citep{chi2020infoxlm}. In addition, we pre-train the model with the IIT-CDIP dataset~\citep{10.1145/1148170.1148307} as well as a great number of publicly available digital-born multilingual PDF files from the internet, which helps the LayoutXLM model to learn from real-world documents. In this way, the model obtains textual and visual signals from a variety of document templates/layouts/formats in different languages, thereby taking advantage of the local invariance property from both textual, visual and linguistic perspectives. Furthermore, to facilitate the evaluation of the pre-trained LayoutXLM model, we employ human annotators to label a multilingual form understanding dataset, which contains 7 languages, including Chinese, Japanese, Spanish, French, Italian, German, Portuguese, and introduces a multilingual benchmark dataset named \textbf{\task} for each language where key-value pairs are annotated. Experiment results show that the pre-trained LayoutXLM outperforms several SOTA cross-lingual pre-trained models on the \task benchmark dataset, which also demonstrates the potential of the multimodal pre-training strategy for multilingual document understanding.

\begin{figure*}[t]
  \centering
  \includegraphics[max width=\textwidth]{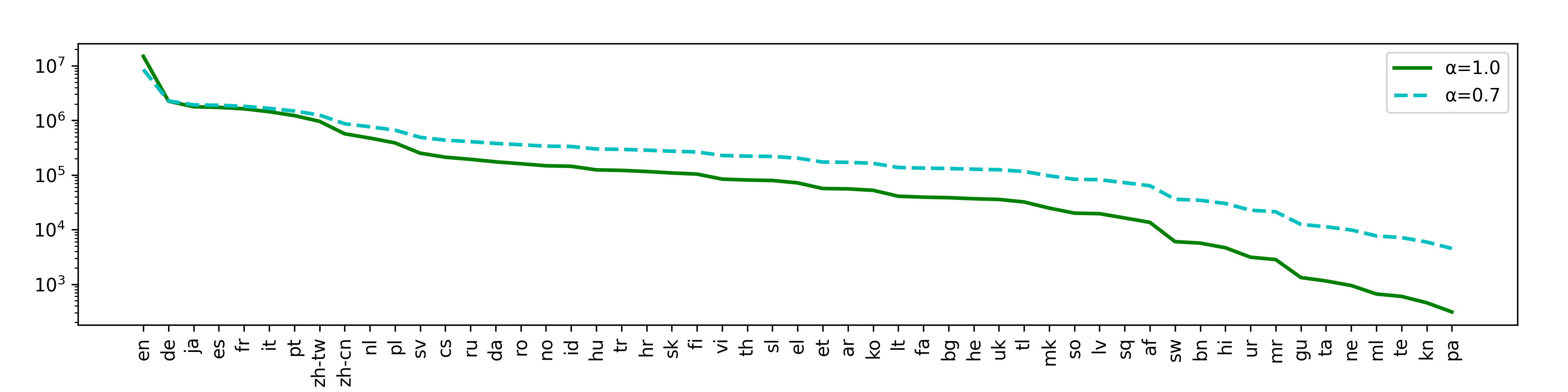}
  \caption{Language distribution of the dataset for pre-training LayoutXLM}
  \label{fig:lang_dist}
\end{figure*}

The contributions of this paper are summarized as follows:

\begin{itemize}
	\item We propose LayoutXLM, a multimodal pre-trained model for multilingual document understanding, which is trained with large-scale real-world scanned/digital-born documents.
	\item We also introduce \task, a multilingual form understanding benchmark dataset that includes human-labeled forms with key-value pairs in 7 languages (Chinese, Japanese, Spanish, French, Italian, German, Portuguese).
	\item LayoutXLM has outperformed other SOTA multilingual baseline models on the \task dataset, which demonstrates the great potential for the multimodal pre-training for the multilingual VrDU task. The pre-trained LayoutXLM model and the \task dataset are publicly available at \url{https://aka.ms/layoutxlm}.
\end{itemize}

% \begin{figure*}[ht]
%     \centering
%     \includegraphics[width=0.95\textwidth]{LayoutLM-v2.pdf}
%     \caption{An illustration of model architectures and pre-training strategies for LayoutLM-v2.}
%     \label{fig:2}
% \end{figure*}

\section{Approach}
In this section, we introduce the model architecture, pre-training objectives, and pre-training dataset. We follow the LayoutLMv2~\cite{xu2020layoutlmv2} architecture and transfer the model to large-scale multilingual document datasets.

\subsection{Model Architecture}
Similar to the LayoutLMv2 framework, we built the LayoutXLM model with a multimodal Transformer architecture. The framework is shown in Figure~\ref{fig:arch}. The model accepts information from three different modalities, including text, layout, and image, which are encoded respectively with text embedding, layout embedding, and visual embedding layers. The text and image embeddings are concatenated, then plus the layout embedding to get the input embedding. The input embeddings are encoded by a multimodal Transformer with the spatial-ware self-attention mechanism. Finally, the output contextual representation can be utilized for the following task-specific layers. For brevity, we refer to \cite{xu2020layoutlmv2} for further details on architecture.

\subsection{Pre-training}
The pre-training objectives of LayoutLMv2 have shown effectiveness in modeling visually-rich documents. Therefore, we naturally adapt this pre-training framework to multilingual document pre-training. Following the idea of cross-modal alignment, our pre-training framework for document understanding contains three pre-training objectives, which are Multilingual Masked Visual-Language Modeling (text-layout alignment), Text-Image Alignment (fine-grained text-image alignment), and Text-Image Matching (coarse-grained text-image alignment).

\paragraph{Multilingual Masked Visual-Language Modeling}
The Masked Visual-Language Modeling (MVLM) is originally proposed in the vanilla LayoutLM and also used in LayoutLMv2, aiming to model the rich text in visually-rich documents. In this pre-training objective, the model is required to predict the masked text token based on its remaining text context and whole layout clues. Similar to the LayoutLM/LayoutLMv2, we train the LayoutXLM with the Multilingual Masked Visual-Language Modeling objective (MMVLM). 

In LayoutLM/LayoutLMv2, an English word is treated as the basic unit, and its layout information is obtained by extracting the bounding box of each word with OCR tools, then subtokens of each word share the same layout information. However, for LayoutXLM, this strategy is not applicable because the definition of the linguistic unit is different from language to language. To prevent the language-specific pre-processing, we decide to obtain the character-level bounding boxes. After the tokenization using SentencePiece with a unigram language model, we calculate the bounding box of each token by merging the bounding boxes of all characters it contains. In this way, we can efficiently unify the multilingual multimodal inputs.

\paragraph{Text-Image Alignment}
The Text-Image Alignment (TIA) task is designed to help the model capture the fine-grained alignment relationship between text and image. We randomly select some text lines and then cover their corresponding image regions on the document image. The model needs to predict a binary label for each token based on whether it is covered or not.

\paragraph{Text-Image Matching}
For Text-Image Matching (TIM), we aim to align the high-level semantic representation between text and image. To this end, we require the model to predict whether the text and image come from the same document page.

\subsection{Pre-training Data}
The LayoutXLM model is pre-trained with documents in 53 languages. Figure~\ref{fig:lang_dist} shows the distribution of pre-training languages. %The complete list of languages is provided in Appendix.
In this section, we briefly describe the pipeline for preparing the large-scale multilingual document collection.

\paragraph{Data Collection}
%LayoutLMv2 uses the IIT-CDIP Test Collection 1.0, which includes nearly 11 million scanned English documents, during the pre-training stage.
% Public webpages and documents on Internet are important sources of high-quality datasets for a variety of languages, including many low-resource ones. Meanwhile, many projects, like Common Crawl\footnote{\url{https://commoncrawl.org}}, have demonstrated the feasibility on pure text. 

To collect a large-scale multilingual visually-rich document collection, we download and process publicly available multilingual digital-born PDF documents following the principles and policies of Common Crawl\footnote{\url{https://commoncrawl.org}}. Using digital-born PDF documents can benefit the collecting and pre-processing steps. On the one hand, we do not have to identify scanned documents among the natural images. On the other hand, we can directly extract accurate text with corresponding layout information with off-the-shelf PDF parsers and save time for running expensive OCR tools. %Finally, we use the URL snapshots from the internet and filter out 680 million multilingual PDF documents.
\paragraph{Pre-processing}
The pre-processing step is needed to clean the dataset since the raw multilingual PDFs are often noisy. We use an open-source PDF parser called PyMuPDF\footnote{\url{https://github.com/pymupdf/PyMuPDF}} to extract text, layout, and document images from PDF documents. After PDF parsing, we discard the documents with less than 200 characters. We use the language detector from the BlingFire\footnote{\url{https://github.com/microsoft/BlingFire}} library and split data per language. Following CCNet~\citep{DBLP:journals/corr/abs-1911-00359}, we classify the document as the corresponding language if the language score is higher than 0.5. Otherwise, unclear PDF files with a language score of less than 0.5 are discarded.

\paragraph{Data Sampling}
After splitting the data per language, we use the same sampling probability $p_l \propto (n_l/n)^{\alpha}$ as XLM~\cite{lample2019crosslingual} to sample the batches from different languages. Following InfoXLM~\citep{chi2020infoxlm}, we use alpha = 0.7 for LayoutXLM to make a reasonable compromise between performance on high- and low-resource languages. The brief language distribution is shown in Figure~\ref{fig:lang_dist}. Finally, we follow this distribution and sample a multilingual document dataset with 22 million visually rich documents. In addition, we also sample 8 million scanned English documents from the IIT-CDIP dataset so that we totally use 30 million documents to pre-train the LayoutXLM, where the model can benefit from the visual information of both scanned and digital-born document images.

\begin{figure*}[t]
	\centering
% 	\begin{subfigure}[b]{0.23\textwidth}
% 		\includegraphics[width=\textwidth]{XFUND.en.jpg}
% 		\caption{English}
% 		\label{fig:1a}
% 	\end{subfigure}
%	~ %add desired spacing between images, e. g. ~, \quad, \qquad, \hfill etc. 
	\begin{subfigure}[b]{0.4\textwidth}
		\includegraphics[width=\textwidth]{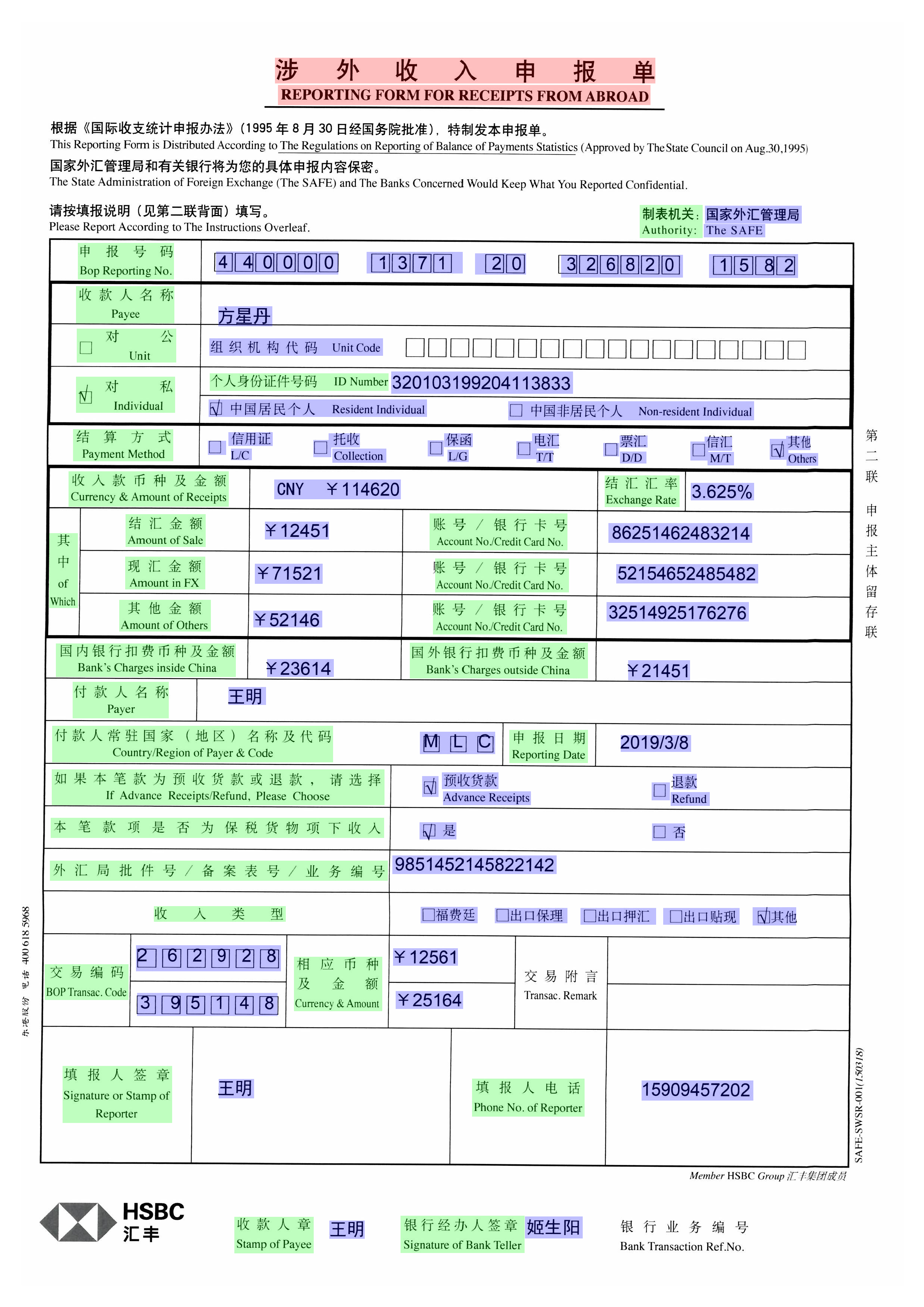}
		\caption{Chinese}
		\label{fig:1b}
	\end{subfigure}
	~ %add desired spacing between images, e. g. ~, \quad, \qquad, \hfill etc. 
% 	\begin{subfigure}[b]{0.23\textwidth}
% 		\includegraphics[width=\textwidth]{XFUND.ja.jpg}
% 		\caption{Japanese}
% 		\label{fig:1c}
% 	\end{subfigure}
% 	~
	\begin{subfigure}[b]{0.4\textwidth}
		\includegraphics[width=\textwidth]{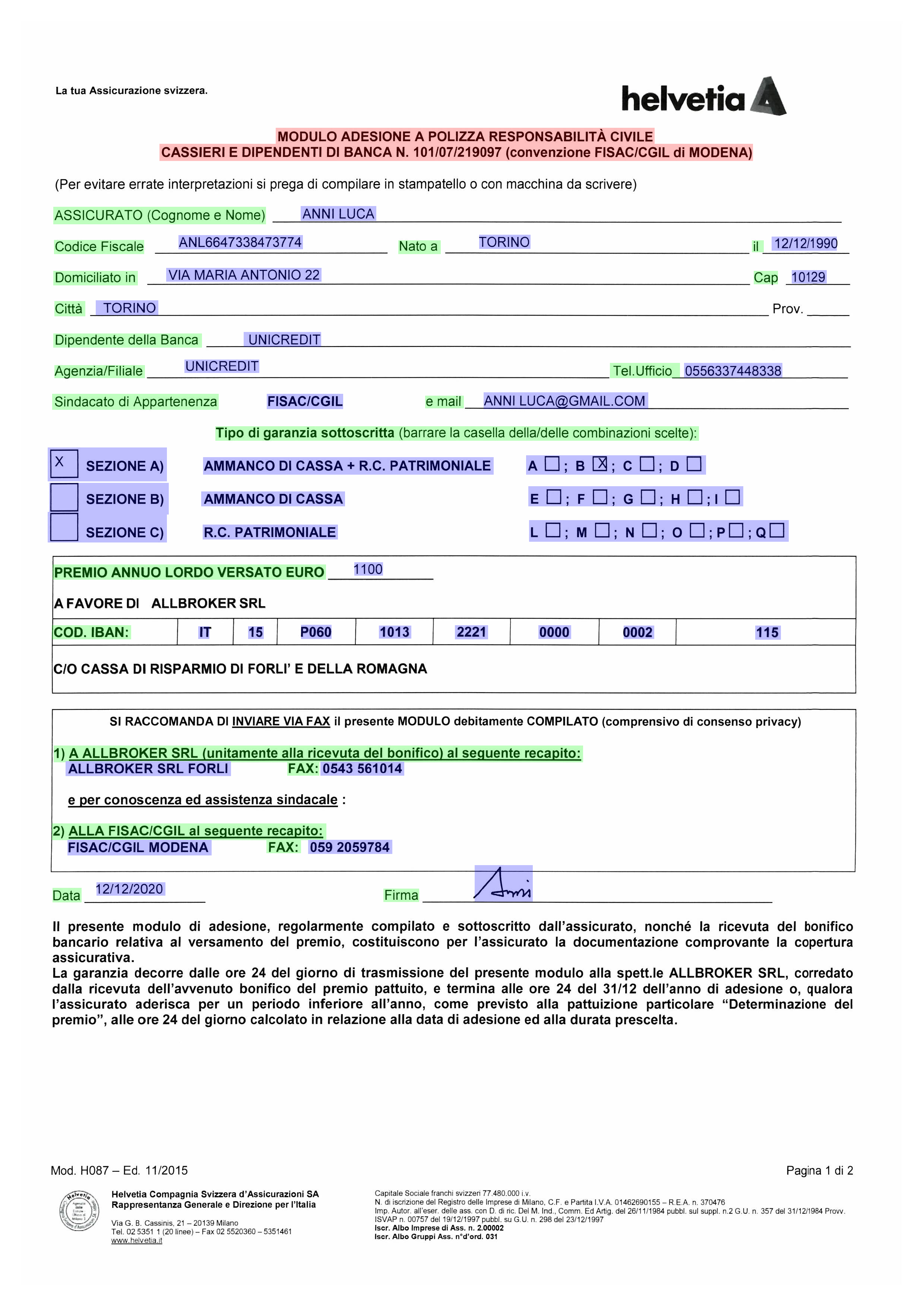}
		\caption{Italian}
		\label{fig:1d}
	\end{subfigure}
% 	~
% 	\newline
% 	\begin{subfigure}[b]{0.23\textwidth}
% 		\includegraphics[width=\textwidth]{XFUND.fr.jpg}
% 		\caption{French}
% 		\label{fig:1e}
% 	\end{subfigure}
% 	~
% 	\begin{subfigure}[b]{0.23\textwidth}
% 		\includegraphics[width=\textwidth]{XFUND.it.jpg}
% 		\caption{Italian}
% 		\label{fig:1f}
% 	\end{subfigure}
% 	~
% 	\begin{subfigure}[b]{0.23\textwidth}
% 		\includegraphics[width=\textwidth]{XFUND.de.jpg}
% 		\caption{German}
% 		\label{fig:1g}
% 	\end{subfigure}
% 	~
% 	\begin{subfigure}[b]{0.23\textwidth}
% 		\includegraphics[width=\textwidth]{XFUND.pt.jpg}
% 		\caption{Portuguese}
% 		\label{fig:1h}
% 	\end{subfigure}
	\caption{Two sampled forms from the \task benchmark dataset (Chinese and Italian), where {\color{red} red} denotes the headers, {\color{green} green} denotes the keys and {\color{blue} blue} denotes the values.}\label{fig:2}
\end{figure*}

\section{\task: A Multilingual Form Understanding Benchmark}

In recent years, many evaluation datasets for document understanding tasks have been proposed, such as PublayNet~\cite{zhong2019publaynet}, FUNSD~\cite{Jaume_2019}, SROIE\footnote{\url{https://rrc.cvc.uab.es/?ch=13}},  TableBank~\cite{li-etal-2020-tablebank}, DocBank~\cite{li-etal-2020-docbank}, DocVQA~\cite{mathew2020docvqa} etc. They have successfully helped to evaluate the proposed neural network models and show the performance gap between the deep learning models and human beings, which significantly empowers the development of document understanding research. However, almost all of these evaluations and benchmarks are solely focused on English documents, which limits the research for non-English document understanding tasks. To this end, we introduce a new benchmark for multilingual Form Understanding, or \task, by extending the FUNSD dataset to 7 other languages, including Chinese, Japanese, Spanish, French, Italian, German, and Portuguese, where sampled documents are shown in Figure~\ref{fig:2}. Next, we introduce the key-value extraction task in our benchmark, as well as data collection, labeling pipeline, and the data statistics. 
% The \task benchmark includes 7 languages with 1,393 fully annotated forms. Each language includes 199 forms, where the training set includes 149 forms, and the test set includes 50 forms.

\subsection{Task description}
Key-value extraction is one of the most critical tasks in form understanding. Similar to FUNSD, we define this task with two sub-tasks, which are semantic entity recognition and relation extraction.
\paragraph{Semantic Entity Recognition}
Given a visually-rich document $\mathcal{D}$, we acquire discrete token set $t=\{t_0, t_1, ..., t_n\}$, where each token $t_i=(w, (x_0, y_0, x_1, y_1))$ consists of a word $w$ and its bounding box coordinates $(x_0, y_0, x_1, y_1)$. $\mathcal{C} = \{c_0, c_1, .., c_m\}$ is the semantic entity labels where the tokens are classified into.
Semantic entity recognition is the task of extracting semantic entities and classifying them into given entity types. In other words, we intend to find a function $F_{SER}:(\mathcal{D},\mathcal{C})\rightarrow \mathcal{E}$, where $\mathcal{E}$ is the predicted semantic entity set:
\begin{equation*}
	\mathcal{E} = \{
	(\{t_0^0, ..., t_0^{n_0} \}, c_0),
	...,
	(\{t_k^0, ..., t_k^{n_k}  \}, c_k)
	\}
\end{equation*}

\paragraph{Relation Extraction}
Equipped with the document $\mathcal{D}$ and the semantic entity label set $\mathcal{C}$, relation extraction aims to predict the relation between any two predicted semantic entities. Defining $\mathcal{R} = \{r_0, r_1, .., r_m\}$ as the semantic relation labels, we intend to find a function $F_{RE}:(\mathcal{D},\mathcal{C},\mathcal{R},\mathcal{E})\rightarrow \mathcal{L}$, where $\mathcal{L}$ is the predicted semantic relation set:
\begin{equation*}
	\mathcal{L} = \{
	(head_0, tail_0, r_0),
	...,
	(head_k, tail_k, r_k)
	\}
\end{equation*}
where $head_i$ and $tail_i$ are two semantic entities. In this work, we mainly focus on the key-value relation extraction.

\subsection{Data Collection and Labeling}

\paragraph{Form Templates}
Forms are usually used to collect information in different business scenarios. In order to avoid the sensitive information leak with the real-world documents, we collect the documents publicly available on the internet and remove the content within the documents while only keeping the templates to manually fill in synthetic information. We collect form templates in 7 languages from the internet. After that, the human annotators manually fill synthetic information into these form templates following corresponding requirements. Each template is allowed to be used only once, which means each form is different from the others. Besides, since the original FUNSD documents contain both digitally filled-out forms and handwritten forms, we also ask annotators to fill in the forms by typing or handwriting. The completed forms are finally scanned into document images for further OCR processing and key-value labeling.

\paragraph{Key-value Pairs}
Key-value pairs are also annotated by human annotators. Equipped with the synthetic forms, we use Microsoft Read API\footnote{\url{https://docs.microsoft.com/en-us/azure/cognitive-services/computer-vision/overview-ocr}} to generate OCR tokens with bounding boxes. With a GUI annotation tool, annotators are shown the original document images and the bounding boxes visualization of all OCR tokens. The annotators are asked to group the discrete tokens into entities and assign pre-defined labels to the entities. Also, if two entities are related, they should be linked together as a key-value pair.

\subsection{Dataset Statistics}
The \task benchmark includes 7 languages with 1,393 fully annotated forms. Each language includes 199 forms, where the training set includes 149 forms, and the test set includes 50 forms. Detailed information is shown in Table~\ref{tab:XFUND}.

\subsection{Baselines}

\paragraph{Semantic Entity Recognition}
For this task, we simply follow the typical sequence labeling paradigm with BIO labeling format and build task-specific layers over the text part of LayoutXLM. 

\paragraph{Relation Extraction}
Following \citet{DBLP:journals/eswa/BekoulisDDD18a} , we first incrementally construct the set of relation candidates by producing all possible pairs of given semantic entities. For every pair, the representation of the head/tail entity is the concatenation of the first token vector in each entity and the entity type embedding obtained with a specific type embedding layer. After respectively projected by two FFN layers, the representations of head and tail are concatenated and then fed into a bi-affine classifier.

\section{Experiments}
In this section, we introduce the experiment settings for pre-training LayoutXLM. To verify the effectiveness of the pre-trained LayoutXLM model, we evaluate all the pre-trained models on our human-labeled \task benchmark.

\subsection{Settings}
\paragraph{Pre-training LayoutXLM}
Following the original LayoutLMv2 recipe, we train LayoutXLM models with two model sizes. For the LayoutXLM\textsubscript{BASE} model, we use a 12-layer Transformer encoder with 12 heads and set the hidden size to $d=768$. For the LayoutXLM\textsubscript{LARGE} model, we increase the layer number to 24 with 16 heads and hidden size to $d=1,024$. ResNeXt101-FPN is used as a visual backbone in both models. Finally, the number of parameters in these two models are approximately 345M and 625M.
During the pre-training stage, we first initialize the Transformer encoder along with text embeddings from InfoXLM and initialize the visual embedding layer with a Mask-RCNN model trained on PubLayNet. The rest of the parameters are initialized randomly. Our models are trained with 64 Nvidia V100 GPUs.

\paragraph{Fine-tuning on \task}
We conduct experiments on the XFUND benchmark. Besides the experiments of typical language-specific fine-tuning, we also design two additional settings to demonstrate the ability to transfer knowledge among different languages, which are zero-shot transfer learning and multitask fine-tuning. Specifically, (1) language-specific fine-tuning refers to the typical fine-tuning paradigm of fine-tuning on language X and testing on language X. (2) Zero-shot transfer learning means the models are trained on English data only and evaluated on each target language. (3) Multitask fine-tuning requires the model to train on data in all languages. We evaluate models in these three settings over two sub-tasks in \task: semantic entity recognition and relation extraction, and compare LayoutXLM to two strong cross-lingual language models: XLM-R and InfoXLM. 

\begin{table}[t]
\small
\centering

\resizebox{0.9\textwidth}{!}{\begin{minipage}{\textwidth}

\begin{tabular}{@{}ccccccc@{}}
\toprule
\bf lang                & \bf split    &\bf header &\bf question &\bf answer &\bf other &\bf total \\ \midrule
\multirow{2}{*}{ZH} & training & 229    & 3,692     & 4,641   & 1,666   & 10,228  \\ \cmidrule(l){2-7} 
                    & testing  & 58    & 1,253     & 1,732    & 586   & 3,629  \\ \midrule
\multirow{2}{*}{JA} & training & 150    & 2,379     & 3,836   & 2,640  & 9,005 \\ \cmidrule(l){2-7} 
                    & testing  & 58     & 723     & 1,280   & 1,322   & 3,383  \\ \midrule
\multirow{2}{*}{ES} & training & 253    & 3,013     & 4,254   & 3,929  & 11,449 \\ \cmidrule(l){2-7} 
                    & testing  & 90     & 909      & 1,218   & 1,196  & 3,413  \\ \midrule
\multirow{2}{*}{FR} & training & 183    & 2,497     & 3,427   & 2,709  & 8,816  \\ \cmidrule(l){2-7} 
                    & testing  & 66     & 1,023     & 1,281   & 1,131  & 3,501  \\ \midrule
\multirow{2}{*}{IT} & training & 166    & 3,762     & 4,932   & 3,355  & 12,215 \\ \cmidrule(l){2-7} 
                    & testing  & 65     & 1,230     & 1,599   & 1,135  & 4,029  \\ \midrule
\multirow{2}{*}{DE} & training & 155    & 2,609     & 3,992   & 1,876  & 8,632  \\ \cmidrule(l){2-7} 
                    & testing  & 59     & 858      & 1,322   & 650   & 2,889  \\ \midrule
\multirow{2}{*}{PT} & training & 185    & 3,510     & 5,428   & 2,531  & 11,654 \\ \cmidrule(l){2-7} 
                    & testing  & 59     & 1,288     & 1,940   & 882   & 4,169 \\ \bottomrule
\end{tabular}
\end{minipage} }
\caption{Statistics of the \task dataset. Each number in the table indicates the number of entities in each category.}
\label{tab:XFUND}
\end{table}

\begin{table*}[ht]
	\small
	\centering
	\begin{tabular}{c|lccccccccc}
		\toprule
		%   \multicolumn{1}{c}{\bf Model} & \bf English & \bf Chinese & \bf Japanese & \bf Spanish & \bf French & \bf Italian & \bf German  & \bf Portuguese \\\midrule
		                    & \multicolumn{1}{c}{\bf Model}      & \bf FUNSD & \bf ZH & \bf JA & \bf ES & \bf FR & \bf IT & \bf DE & \bf PT & \bf Avg. \\\midrule
		\multirow{6}{*}{SER} & $\textrm{XLM-RoBERTa}_{\rm BASE}$  & 0.667  & 0.8774 & 0.7761 & 0.6105 & 0.6743 & 0.6687 & 0.6814 & 0.6818 & 0.7047      \\
		                    & $\textrm{InfoXLM}_{\rm BASE}$      & 0.6852 & 0.8868 & 0.7865 & 0.6230 & 0.7015 & 0.6751 & 0.7063 & 0.7008 & 0.7207      \\
		                    & $\textrm{LayoutXLM}_{\rm BASE}$    & \bf 0.794  & \bf 0.8924 & \bf 0.7921 & \bf 0.7550 & \bf 0.7902 & \bf 0.8082 & \bf 0.8222 & \bf 0.7903 & \bf 0.8056      \\ \cmidrule{2-11}
		                    & $\textrm{XLM-RoBERTa}_{\rm LARGE}$ & 0.7074 & 0.8925 & 0.7817 & 0.6515 & 0.7170 & 0.7139 & 0.711  & 0.7241 & 0.7374      \\
		                    & $\textrm{InfoXLM}_{\rm LARGE}$     & 0.7325 & 0.8955 & 0.7904 & 0.6740 & 0.7140 & 0.7152 & 0.7338 & 0.7212 & 0.7471      \\
		                    & $\textrm{LayoutXLM}_{\rm LARGE}$   & \bf 0.8225 & \bf 0.9161 & \bf 0.8033 & \bf 0.7830 & \bf 0.8098 & \bf 0.8275 & \bf 0.8361 & \bf 0.8273 & \bf 0.8282      \\\midrule
		\multirow{6}{*}{RE} & $\textrm{XLM-RoBERTa}_{\rm BASE}$  & 0.2659 & 0.5105 & 0.5800 & 0.5295 & 0.4965 & 0.5305 & 0.5041 & 0.3982 & 0.4769 \\
		                    & $\textrm{InfoXLM}_{\rm BASE}$      & 0.2920 & 0.5214 & 0.6000 & 0.5516 & 0.4913 & 0.5281 & 0.5262 & 0.4170 & 0.4910 \\
		                     & $\textrm{LayoutXLM}_{\rm BASE}$    & \bf 0.5483 & \bf 0.7073 & \bf 0.6963 & \bf 0.6896 & \bf 0.6353 & \bf 0.6415 & \bf 0.6551 & \bf 0.5718 & \bf 0.6432 \\ \cmidrule{2-11}
		                    & $\textrm{XLM-RoBERTa}_{\rm LARGE}$ & 0.3473 & 0.6475 & 0.6798 & 0.6330 & 0.6080 & 0.6171 & 0.6189 & 0.5762 & 0.5910 \\
		                    & $\textrm{InfoXLM}_{\rm LARGE}$     & 0.3679 & 0.6775 & 0.6604 & 0.6346 & 0.6096 & 0.6659 & 0.6057 & 0.5800 & 0.6002 \\
		                    & $\textrm{LayoutXLM}_{\rm LARGE}$   & \bf 0.6404 & \bf 0.7888 & \bf 0.7255 & \bf 0.7666 & \bf 0.7102 & \bf 0.7691 & \bf 0.6843 & \bf 0.6796 & \bf 0.7206 \\
		\bottomrule
	\end{tabular}
	\caption{Language-specific fine-tuning accuracy (F1) on the \task dataset (fine-tuning on X, testing on X), where ``SER'' denotes the semantic entity recognition and ``RE'' denotes the relation extraction.}
	\label{tab:x2x}
\end{table*}

\begin{table*}[ht]
	\small
	\centering
	\begin{tabular}{c|lccccccccc}
		\toprule
		%   \multicolumn{1}{c}{\bf Model} & \bf English & \bf Chinese & \bf Japanese & \bf Spanish & \bf French & \bf Italian & \bf German  & \bf Portuguese \\\midrule
		                    & \multicolumn{1}{c}{\bf Model}      & \bf FUNSD & \bf ZH & \bf JA & \bf ES & \bf FR & \bf IT & \bf DE & \bf PT & \bf Avg. \\\midrule
		\multirow{6}{*}{SER} & $\textrm{XLM-RoBERTa}_{\rm BASE}$  & 0.667  & 0.4144 & 0.3023 & 0.3055 & 0.371  & 0.2767 & 0.3286 & 0.3936 & 0.3824 \\
		                    & $\textrm{InfoXLM}_{\rm BASE}$ & 0.6852 & 0.4408 & 0.3603 & 0.3102 & 0.4021 & 0.2880 & 0.3587 & 0.4502 & 0.4119 \\
		                    & $\textrm{LayoutXLM}_{\rm BASE}$    & \bf 0.794  & \bf 0.6019 & \bf 0.4715 & \bf 0.4565 & \bf 0.5757 & \bf 0.4846 & \bf 0.5252 & \bf 0.539 & \bf 0.5561 \\ \cmidrule{2-11}
		                    & $\textrm{XLM-RoBERTa}_{\rm LARGE}$ & 0.7074 & 0.5205 & 0.3939 & 0.3627 & 0.4672 & 0.3398 & 0.418  & 0.4997 & 0.4637 \\
		                    & $\textrm{InfoXLM}_{\rm LARGE}$     & 0.7325 & 0.5536 & 0.4132 & 0.3689 & 0.4909 & 0.3598 & 0.4363 & 0.5126 & 0.4835 \\
		                    & $\textrm{LayoutXLM}_{\rm LARGE}$   & \bf 0.8225 & \bf 0.6896 & \bf 0.519  & \bf 0.4976 & \bf 0.6135 & \bf 0.5517 & \bf 0.5905 & \bf 0.6077 & \bf 0.6115 \\\midrule
		\multirow{6}{*}{RE} & $\textrm{XLM-RoBERTa}_{\rm BASE}$  & 0.2659 & 0.1601 & 0.2611 & 0.2440 & 0.2240 & 0.2374 & 0.2288 & 0.1996 & 0.2276 \\
		                    & $\textrm{InfoXLM}_{\rm BASE}$      & 0.2920 & 0.2405 & 0.2851 & 0.2481 & 0.2454 & 0.2193 & 0.2027 & 0.2049 & 0.2423 \\
		                    & $\textrm{LayoutXLM}_{\rm BASE}$    & \bf 0.5483 & \bf 0.4494 & \bf 0.4408 & \bf 0.4708 & \bf 0.4416 & \bf 0.4090 & \bf 0.3820 & \bf 0.3685 & \bf 0.4388 \\ \cmidrule{2-11}
		                    & $\textrm{XLM-RoBERTa}_{\rm LARGE}$ & 0.3473 & 0.2421 & 0.3037 & 0.2843 & 0.2897 & 0.2496 & 0.2617 & 0.2333 & 0.2765 \\
		                    & $\textrm{InfoXLM}_{\rm LARGE}$     & 0.3679 & 0.3156 & 0.3364 & 0.3185 & 0.3189 & 0.2720 & 0.2953 & 0.2554 & 0.3100 \\
		                    & $\textrm{LayoutXLM}_{\rm LARGE}$   & \bf 0.6404 & \bf 0.5531 & \bf 0.5696 & \bf 0.5780 & \bf 0.5615 & \bf 0.5184 & \bf 0.4890 & \bf 0.4795 \bf & \bf 0.5487 \\
		\bottomrule
	\end{tabular}
	\caption{Zero-shot transfer accuracy (F1) on the \task dataset (fine-tuning on FUNSD, testing on X), where ``SER'' denotes the semantic entity recognition and ``RE'' denotes the relation extraction.}
	\label{tab:e2x}
\end{table*}

\begin{table*}[ht]
	\small
	\centering
	\begin{tabular}{c|lccccccccc}
		\toprule
		%   \multicolumn{1}{c}{\bf Model} & \bf English & \bf Chinese & \bf Japanese & \bf Spanish & \bf French & \bf Italian & \bf German  & \bf Portuguese \\\midrule
		                    & \multicolumn{1}{c}{\bf Model}      & \bf FUNSD & \bf ZH & \bf JA & \bf ES & \bf FR & \bf IT & \bf DE & \bf PT & \bf Avg. \\\midrule
		\multirow{6}{*}{SER} & $\textrm{XLM-RoBERTa}_{\rm BASE}$  & 0.6633 & 0.883  & 0.7786 & 0.6223 & 0.7035 & 0.6814 & 0.7146 & 0.6726 & 0.7149 \\
		                    & $\textrm{InfoXLM}_{\rm BASE}$      & 0.6538 & 0.8741 & 0.7855 & 0.5979 & 0.7057 & 0.6826 & 0.7055 & 0.6796 & 0.7106 \\
		                    & $\textrm{LayoutXLM}_{\rm BASE}$    & \bf 0.7924 & \bf 0.8973 & \bf 0.7964 & \bf 0.7798 & \bf 0.8173 & \bf 0.821  & \bf 0.8322 & \bf 0.8241 & \bf 0.8201 \\ \cmidrule{2-11}
		                    & $\textrm{XLM-RoBERTa}_{\rm LARGE}$ & 0.7151 & 0.8967 & 0.7828 & 0.6615 & 0.7407 & 0.7165 & 0.7431 & 0.7449 & 0.7502 \\
		                    & $\textrm{InfoXLM}_{\rm LARGE}$     & 0.7246 & 0.8919 & 0.7998 & 0.6702 & 0.7376 & 0.7180 & 0.7523 & 0.7332 & 0.7534 \\
		                    & $\textrm{LayoutXLM}_{\rm LARGE}$   & \bf 0.8068 & \bf 0.9155 & \bf 0.8216 & \bf 0.8055 & \bf 0.8384 & \bf 0.8372 & \bf 0.853  & \bf 0.8650 & \bf 0.8429 \\\midrule
		\multirow{6}{*}{RE} & $\textrm{XLM-RoBERTa}_{\rm BASE}$  &0.3638 & 0.6797 & 0.6829 & 0.6828 & 0.6727 & 0.6937 & 0.6887 & 0.6082 & 0.6341 \\
		                    & $\textrm{InfoXLM}_{\rm BASE}$      & 0.3699 & 0.6493 & 0.6473 & 0.6828 & 0.6831 & 0.6690 & 0.6384 & 0.5763 & 0.6145\\
		                    & $\textrm{LayoutXLM}_{\rm BASE}$    & \bf 0.6671 & \bf 0.8241 & \bf 0.8142 & \bf 0.8104 & \bf 0.8221 & \bf 0.8310 & \bf 0.7854 & \bf 0.7044 & \bf 0.7823\\ \cmidrule{2-11}
		                    & $\textrm{XLM-RoBERTa}_{\rm LARGE}$ &0.4246 & 0.7316 & 0.7350 & 0.7513 & 0.7532 & 0.7520 & 0.7111 & 0.6582 & 0.6896\\
		                    & $\textrm{InfoXLM}_{\rm LARGE}$     & 0.4543 & 0.7311 & 0.7510 & 0.7644 & 0.7549 & 0.7504 & 0.7356 & 0.6875 & 0.7037 \\
		                    & $\textrm{LayoutXLM}_{\rm LARGE}$   & \bf 0.7683 & \bf 0.9000 & \bf 0.8621 & \bf 0.8592 & \bf 0.8669 & \bf 0.8675 & \bf 0.8263 & \bf 0.8160 & \bf 0.8458 \\
		\bottomrule
	\end{tabular}
	\caption{Multitask fine-tuning accuracy (F1) on the \task dataset (fine-tuning on 8 languages all, testing on X), where ``SER'' denotes the semantic entity recognition and ``RE'' denotes the relation extraction.}
	\label{tab:all2x}
\end{table*}

\subsection{Results}
We evaluate the LayoutXLM model on language-specific fine-tuning tasks, and the results are shown in Table ~\ref{tab:x2x}. Compared with the pre-trained models such as XLM-R and InfoXLM, the LayoutXLM LARGE model achieves the highest F1 scores in both SER and RE tasks. The significant improvement shows LayoutXLM's capability to transfer knowledge obtained from pre-training to downstream tasks, which further confirms the effectiveness of our multilingual pre-training framework.

For the cross-lingual zero-shot transfer, we present the evaluation results in Table ~\ref{tab:e2x}. Although the model are only fine-tuned on FUNSD dataset (in English), it can still transfer the knowledge to different languages. In addition, it is observed that the LayoutXLM model significantly outperforms the other text-based models. This verifies that LayoutXLM can capture the common layout invariance among different languages and transfer to other languages for form understanding.

Finally, Table ~\ref{tab:all2x} shows the evaluation results on the multitask learning. In this setting, the pre-trained LayoutXLM model is fine-tuned with all 8 languages simultaneously and evaluated on each specific language, in order to investigate whether improvements can be obtained by multilingual fine-tuning. We observe that the multitask learning further improves the model performance compared to the language-specific fine-tuning, which also confirms that document understanding can benefit from the layout invariance among different languages.

% \begin{table*}[ht]
% %\small
%     \centering
%     \begin{tabular}{lcccccccc}
%     \toprule
%     %   \multicolumn{1}{c}{\bf Model} & \bf English & \bf Chinese & \bf Japanese & \bf Spanish & \bf French & \bf Italian & \bf German  & \bf Portuguese \\\midrule
%     \multicolumn{1}{c}{\bf Model} & \bf EN & \bf ZH & \bf JA & \bf ES & \bf FR & \bf IT & \bf DE  & \bf PT \\\midrule
%       $\textrm{XLM-RoBERTa}_{\rm BASE}$  & - & - & - & - \\
%       $\textrm{InfoXLM}_{\rm BASE}$  & - & - & - & -  \\
%       $\textrm{XLM-RoBERTa}_{\rm LARGE}$   & - & - & - & - \\
%       $\textrm{InfoXLM}_{\rm LARGE}$  & - & - & - & - \\\midrule
%       $\textrm{LayoutXLM}_{\rm BASE}$ & - & - & - & - \\
%       $\textrm{LayoutXLM}_{\rm LARGE}$ & - & - & - & - \\
%      \bottomrule
%     \end{tabular}
%     \caption{End-to-end key-value extraction accuracy (F1) on the XFUND dataset (fine-tuning on 8 languages all, testing on X)}
%     \label{tab:kv-all2x}
% \end{table*}

\section{Related Work}

Multimodal pre-training has become popular in recent years due to its successful applications in vision-language representation learning.~\citet{lu2019vilbert} proposed ViLBERT for learning task-agnostic joint representations of image content and natural language by extending the popular BERT architecture to a multimodal two-stream model.~\citet{su2020vlbert} proposed VL-BERT that adopts the Transformer model as the backbone, and extends it to take both visual and linguistic embedded features as input.~\citet{li2019visualbert} propose VisualBERT consists of a stack of Transformer layers that implicitly align elements of an input text and regions in an associated input image with self-attention.~\citet{chen2020uniter} introduced UNITER that learns through large-scale pre-training over four image-text datasets (COCO, Visual Genome, Conceptual Captions, and SBU Captions), which can power heterogeneous downstream V+L tasks with joint multimodal embeddings.~\citet{li2020oscar} proposed a new learning method Oscar (Object-Semantics Aligned Pre-training), which uses object tags detected in images as anchor points to significantly ease the learning of alignments. Inspired by these vision-language pre-trained models, we would like to introduce the vision-language pre-training into the document intelligence area, where the text, layout, and image information can be jointly learned to benefit the VrDU tasks.

Multilingual pre-trained models have pushed many SOTA results on cross-lingual natural language understanding tasks by pre-training the Transformer models on different languages. These models have successfully bridged the language barriers in a number of cross-lingual transfer benchmarks such as XNLI~\citep{conneau2018xnli} and XTREME~\citep{hu2020xtreme}.~\citet{devlin2018bert} introduced a new language representation model called BERT and extend to a multilingual version called mBERT, which is designed to pre-train deep bidirectional representations from the unlabeled text by jointly conditioning on both left and right context in all layers. As a result, the pre-trained BERT model can be fine-tuned with just one additional output layer to create state-of-the-art models for a wide range of tasks.~\citet{lample2019crosslingual} proposed two methods to learn cross-lingual language models (XLMs): one unsupervised that only relies on monolingual data, and one supervised that leverages parallel data with a new cross-lingual language model objective.~\citet{conneau2020unsupervised} proposed to train a Transformer-based masked language model on one hundred languages, using more than two terabytes of filtered CommonCrawl data, which significantly outperforms mBERT on a variety of cross-lingual benchmarks. Recently,~\citet{chi2020infoxlm} formulated cross-lingual language model pre-training as maximizing mutual information between multilingual-multi-granularity texts. The unified view helps to better understand the existing methods for learning cross-lingual representations, and the information-theoretic framework inspires to propose a pre-training task based on contrastive learning.~\citet{liu2020multilingual} presented mBART -- a sequence-to-sequence denoising auto-encoder pre-trained on large-scale monolingual corpora in many languages using the BART objective.~\citet{xue2020mt5} introduced mT5, a multilingual variant of T5 that was pre-trained on a new Common Crawl-based dataset covering 101 languages. The pre-trained LayoutXLM model is built on the multilingual
textual models as the initialization, which benefits the VrDU tasks in different languages worldwide.

\section{Conclusion}

In this paper, we present LayoutXLM, a multimodal pre-trained model for multilingual visually-rich document understanding. 
%By taking advantage of the LayoutLMv2 architecture, 
The LayoutXLM model is pre-trained with 30 million scanned and digital-born documents in 53 languages.
%To effectively evaluate the multilingual pre-trained models,
Meanwhile, we also introduce the multilingual form understanding benchmark \task, which includes key-value labeled forms in 7 languages. Experimental results have illustrated that the pre-trained LayoutXLM model has significantly outperformed
the SOTA baselines for multilingual document understanding, which bridges the language gap in real-world document understanding tasks. We make LayoutXLM and \task publicly available to advance the document understanding research.

For future research, we will further enlarge the multilingual training data to cover more languages as well as more document layouts and templates. In addition, as there are a great number of business documents with the same content but in different languages, we will also investigate how to leverage the contrastive learning of parallel documents for the multilingual pre-training.

\bibliography{anthology,custom}
\bibliographystyle{acl_natbib}

% \appendix

% \section{Example Appendix}
% \label{sec:appendix}

% This is an appendix.

\end{document}